# An Efficient Method for Recognizing the Low Quality Fingerprint Verification by Means of Cross Correlation

V.Karthikeyan[1] and V.J.Vijayalakshmi[2]

[1]Department of ECE, SVSCE, Coimbatore, India
[2]Department of EEE, SKCET, Coimbatore, India

## ABSTRACT

*In this paper, we propose an efficient method to provide personal identification using fingerprint to get better accuracy even in noisy condition. The fingerprint matching based on the number of corresponding minutia pairings, has been in use for a long time, which is not very efficient for recognizing the low quality fingerprints. To overcome this problem, correlation technique is used. The correlation-based fingerprint verification system is capable of dealing with low quality images from which no minutiae can be extracted reliably and with fingerprints that suffer from non-uniform shape distortions, also in case of damaged and partial images. Orientation Field Methodology (OFM) has been used as a preprocessing module, and it converts the images into a field pattern based on the direction of the ridges, loops and bifurcations in the image of a fingerprint. The input image is then Cross Correlated (CC) with all the images in the cluster and the highest correlated image is taken as the output. The result gives a good recognition rate, as the proposed scheme uses Cross Correlation of Field Orientation (CCFO = OFM + CC) for fingerprint identification.*

## 1. INTRODUCTION

Conventional security systems used either knowledge based methods (passwords or PIN), and token-based methods (passport, driver license, ID card) and were prone to fraud because PIN numbers could be forgotten or hacked and the tokens could be lost, duplicated or stolen [7]. Accurate and automatic identification and authentication of users is a fundamental problem in today's computing world [8]. In the last few years, biometric authentication has become an increasingly important issue in modern society. The biometrics are enhancing our ability to identify people. There are two types of biometric techniques: 1. Physiological (face recognition, iris recognition, finger print recognition, retina recognition). 2. Behavioral (signature recognition, keystroke recognition and voice recognition). There are various biometric identification techniques such as palm print [9], fingerprint [10], face [11], vein [12] or their combinations [13]. Among all the biometric techniques, today fingerprints are the most widely used biometric features for personal identification because of their high acceptability, immutability and individuality [14]. Fingerprint verification is one of the most reliable and personal identification methods [1]. Fingerprint images are widely used in many systems such as personal identification, access control, internet authentication, forensics, e-banking, etc. Due to its permanence, uniqueness and distinctiveness [2] In the Table I various biometric technologies have been compared based on various characteristics.

       

TABLE I
Comparison of Various Biometric Technologies

| Biometric Identifier | Um | Di | Pm | Co | Pf | Ac | Ci |
|---|---|---|---|---|---|---|---|
| Face | H | L | M | H | L | H | H |
| Fingerprint | M | H | H | M | H | M | M |
| Hand Geometry | M | M | M | H | M | M | M |
| Iris | H | H | H | M | H | L | L |
| Keystroke | L | L | L | M | L | M | M |
| Signature | L | L | L | M | L | H | H |
| Voice | M | L | L | M | L | H | H |

Un- Universality  Pf – Performance  Di– Distinct
Ci– Circumvention  Pm– Permanence  L – Low
Co – Collectability  M – Medium  H- High

Usually, fingerprint verification is performed manually by professional forensic experts. However, manual fingerprint verification is very tedious. Hence, Automatic Fingerprint Identification Systems (AFIS) are in great demand. There are a number of design factors like lack of reliable minutiae extraction algorithms, difficulty in quantitatively defining a reliable match between fingerprint images, fingerprint classification, etc. creates bottlenecks in achieving the desired performance [3]. Fingerprint has been widely used for personal identification for several centuries [4]. Minutiae extraction - based fingerprint identification is a popular method. But the cross correlation based technique is a promising approach to fingerprint authentication for the new generation of high resolution and touch less fingerprint sensors. This paper proposes a novel scheme, namely, Cross Correlation of Field Orientation (CCFO) that cascades Cross Correlation technique with Field Orientation technique to do fingerprint authentication. This paper is organized as follows: Section II describes about the proposed system including the pre-processing, field orientation estimation and matching modules. In Section III some experimental results are presented. Finally, the conclusions are discussed in Section IV.

## 2. PROPOSED SYSTEM ARCHITECTURE

The overall architecture of the proposed biometric identification system is illustrated in Figure 1. Each of the constituent modules are described in this section, When compared to the feature extraction method the Cross Correlation of Field Orientation method has several features that accounts for its improved performance of fingerprint authentication. Using the OFM the images are converted into field orientation images that which increase the immunity to noise. Cross correlation of images used for matching is a very simple and accurate method for measuring image similarities [6]. Rao's algorithm is used for measuring the field orientation [5]. The following steps are involved in the Rao's algorithm. The image is passed through a low pass filter which smoothes the image. The low pass filter used is median filter. The Gradient of the smoothened image is calculated for x and y axis. The second order gradients are calculated. The resultant field orientation is then divided into $N \times N$ pixel and the orientation is represented by arrows for each block. Then the template is cross correlated with the input image for matching. Theme based on the cross correlation value the decision is made. Some of the pre-processing steps have to be done before getting the Field Orientation image. These steps are done for getting better and accurate results. There are two steps involved in pre-processing. They are smooth and edge detection. Smoothing is a technique used to reduce the noise within an image. It is an important step in image processing. It would be difficult to process the high frequency images which are due to the drastic variation in the pixel intensity. Thus smoothing is done before field orientation to reduce the variations in the pixel intensity. The median filter is used for smoothing in this proposed method. The median filter which is a nonlinear filter is an effective method that can suppress isolated noise without blurring sharp edges. It helps to remove the impulse noise from the image, while preserving the rapid intensity changes. Specifically, the median filter replaces a pixel value at the center of the median of all pixel value in the neighborhood. Median filter is a more robust method than the traditional linear filtering, because it preserves the sharp edges while removing the noise.

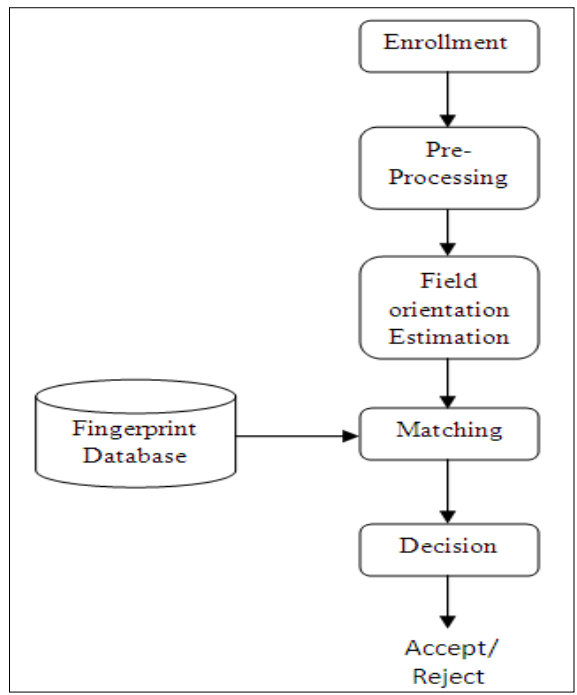

Figure 1. System Architecture

The median filter in 1-d works as, it just sorts the value and considers the middle value as median. The 2-d median filter is illustrated as below

$$G (x, y) = median \{a (I, j), (I, j) \in w\} \qquad (1)$$

Where *w* represents a neighborhood centered around location *(x, y)* in the image and *x* and *y* are the random variables representing the variations along two directions. Edge detection is one of the most commonly used operations in image analysis. It is a fundamental tool used in most image processing applications to obtain information from the frames as a precursor step to feature extraction and feature detection. It refers to the process of identifying and locating sharp discontinuities in an image. The edges form the outline of an object. An edge is the boundary between an object and the background. This process detects outlines of an object and boundaries between objects and the background in the image. Thus the result of applying an edge detector to an image may lead to a set of connected curves that indicate the boundaries of surface markings as well as curves that correspond to discontinuities in surface orientation. Thus, applying an edge detection algorithm to an image may significantly reduce the amount of data to be processed and may therefore filter out information that may be regarded as less relevant, while preserving the important structural properties of an image. Canny filter is extensively used for edge detection**.**

The Canny edge detection algorithm is one of the best optimal edge detectors. The advantage is that it has a low error rate. The canny edge detector finds the image gradient to highlight regions with high spatial derivatives. The gradient of the smoothened image is calculated for x, y axis. Let it be $G_x$ in x direction and $G_y$ in Y direction. The second order gradients are calculated using the following equations

$$G_{xx} = G_x * G_x^T \quad (2)$$
$$G_{xy} = G_x * G_y^T \quad (3)$$
$$G_{yy} = G_y * G_y^T \quad (4)$$

Where $G_{xx}$, $G_{xy}$, $G_{yy}$ are the second order gradients of $G_x$ and $G_y$. $G_x^T$, $G_y^T$ are the transpose matrices of $G_x$ and $G_y$ respectively.

## 3. FIELD ORIENTATION ESTIMATION

Field orientation of a fingerprint image is an efficient technique used to extract the directional properties of the image and not the actual image. However, the gradients are orientations at pixel scale, while the orientation field describes the orientation of the ridge valley structures. Therefore, the field orientation can be derived from the gradients by performing some operation on the gradients. The field orientation is calculated using the following equations.

$$x = \frac{Gxy}{\sqrt{(Gxy2 + (Gxx - Gyy)2)}} \quad (5)$$

$$y = \frac{(Gxx - Gyy)}{\sqrt{(Gxy2 + (Gxx - Gyy)2)}} \quad (6)$$

$$\Theta = \pi/2 + \tan^{-1}(x) + \tan^{-1}(y) \quad (7)$$

The resultant field orientation is then divided into N×N pixel and the orientation is represented by arrows for each block. The Cross Correlation is a technique that which used to determine the degree of similarity between two similar images. The Cross Correlation computation of Template (T) and Input (I) image is determined by the following equation.

$$CC\ (T, I) = \sum_{i=0}^{n-1} \sum_{j=0}^{m-1} T(i,j) I(i,j) \quad (8)$$

Here both T and I represent the field orientation images. In the frequency domain the cross correlation can be calculated by using the following formula.

$$CC\ (T, I) = IFT\ (F'\ (T) * F\ (I)) \quad (9)$$

Here IFT represents the Inverse Fourier Transform. When compared to the Time domain computation, the Fourier domain is more efficient. Here, F' (T) -Complex conjugate of the Fourier Transform of the Template image. F (I) - the Fourier Transform of Input image The cross correlation value is normalized by dividing the value obtained by total number of pixels.

## 4. EXPERIMENTAL RESULTS

The efficiency of the proposed approach is tested on well known MIT database. This database contains the low resolution fingerprints. We used 1000 fingerprints of 200 classes for the experiment. The query image is obtained from fingerprint reader and then it is converted into a grayscale image as shown in Figure 2. From the query image the field orientation image is calculated using Eq.2 to Eq.7. The median filter is used for smoothing and the canny filter is used for edge detection and the results are shown in Figure 3. The Cross Correlation is used for the matching which represents the degree of similarity between the images. Figure 4 shows the result for the authenticated person where the right hand image is the retrieved image and the left hand side image is the input image. The percentage of matching is 85%. When any new image is given as input, it is identified as the intruder which is shown in Figure 5.

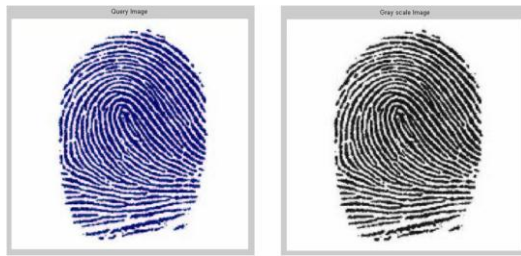

Figure 2. Query image and gray scale image

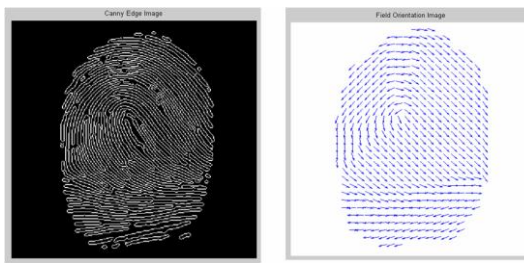

Figure 3. Edge detected image and Field Oriented Image

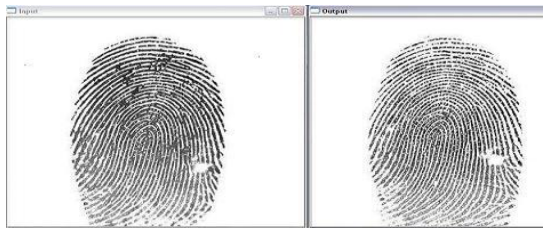

Figure.4. Final output for authenticated person

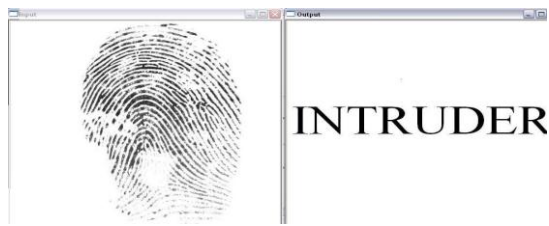

Figure.5. Final output for an Intruder

In the above figure the image which is to be authenticated is not presented in the database. Thus the person is declared as an intruder when the percentage of matching is less than 85%. Figure 6 shows the performance evaluation of the algorithm which is implemented. As the size of the database increases the percentage of the matching gradually decreases. The percentage of matching remains at 85%, for a set of 700 fingerprint images.

## 5. CONCLUSION

In our proposed system we have implemented the Cross Correlation technique in the matching stage. Correlation based techniques are a promising approach to fingerprint authentication. And median filter is used for the image smoothing and the canny filter is used for the edge detection. And median filter is a more robust method than traditional linear filtering. Traditional techniques, like minutiae-based techniques, do not exploit all the information of the low resolution and damaged images. But with the proposed system even with the damaged or partial fingerprint image it is possible to check that a person is authenticated or an intruder.

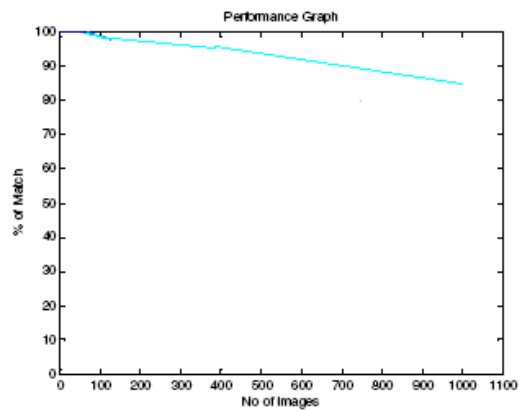

Figure.6. Performance graph

## REFERENCES

[1] Henry C. Lee and R. E. Gaensslen, editors, Advances in Fingerprint Technology, Elsevier, New York, 1991.
[2] D. Maltoni, D. Maio, A. K. Jain, and S. Prabhakar, Handbook of Fingerprint Recognition. Springer, 2003.
[3] Shlomo Greenberg, Mayer Aladjem, Daniel Kogan and Itshak Dimitrov, "Fingerprint Image Enhancement using Filtering Techniques", 15th International conference on Pattern recognition, 3-7 September 2000.
[4] Almudena Lindoso, Luis Entrena, Judith Liu-Jimneez, Enrique San Milan, "Increasing security with correlation based fingerprinting", 41st annual IEEE International Carnahan Conference on security technology, 8-11 October 2007.
[5] Anil Jain , Lin Hong, and Ruud Bolle, "On-line fingerprint verification", IEEE transactions on pattern analysis and machine intelligence, Vol. 19, no. 4, April 1997.
[6] Abhishek Nagar, Karthik Nandakumar, Anil K. Jain: Securing fingerprint template: Fuzzy vault with minutiae descriptors. International Conference on Pattern Recognition 2008: pp.1-4 Award winning papers from the 19th International Conference on Pattern Recognition (ICPR),
[7] Megha Kulshrestha, Pooja, V. K. Banga, "Selection of an Optimal Algorithm for Fingerprint Matching" World Academy of Science, Engineering and Technology 75 2011
[8] Rajeswari Mukesh, Dr. A. Damodaram, Dr. V. Subbiah Bharathi, "A Robust Finger Print based Two-Server Authentication and Key Exchange System".

[9] Kong A., Zhang D. And Mohamed K., "Three measures for secure palmprint identification", Pattern Recognition, 2008, 41, pp.1329-1337.
[10] M. Zsolt, K.V., "A Fingerprint Verification System Based on Triangular Matching and Dynamic Time Warping", IEEE transactions on Pattern Analysis and Machine Intelligence, 2000, 22(11), pp.1266-1276.
[11] Andrew B.J., Alwyn G. and David C.L., "Random Multispace Quantization as an Analytic Mechanism for BioHashing of Biometric and Random Identity Inputs", IEEE transactions on Pattern Analysis and Machine Intelligence, 2006, 28 (12), pp. 1892-1901.
[12] Lingyu Wang , Graham Leedham , David Siu-Yeung Cho, "Minutiae feature analysis for infrared hand vein pattern biometrics", Pattern Recognition, 2008, 41 (3), pp. 920-929.
[13] Kumar A., Zhang D., "Combining fingerprint, palm print and hand-shape for user authentication", The 18th International Conference on Pattern Recognition (ICPR'06), 2006, 4, pp.549-552
[14] Vaidehi. V , Naresh Babu N T, Ponsamuel Mervin.A, Praveen Kumar.S, Velmurugan.S, Balamurali, Girish Chandra, "Fingerprint Identification Using Cross Correlation of Field Orientation", ICoAC 2010 IEEE.

**Authors**

**Prof.V.Karthikeyan** has received his Bachelor's Degree in Electronics and Communication Engineering from PGP college of Engineering and Technology in 2003, Namakkal, India, He received a Masters Degree in Applied Electronics from KSR college of Technology, Erode in 2006 He is currently working as Assistant Professor in SVS College of Engineering and Technology, Coimbatore. He has about 8 years of Teaching Experience 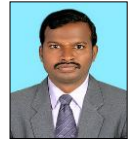

**Prof. V. J. Vijayalakshmi** has completed her Bachelor's Degree Electrical & Electronics Engineering from Sri Ramakrishna Engineering College, Coimbatore, India. She finished her Masters Degree in Power Systems Engineering from Anna University of Technology, Coimbatore, She is currently working as Assistant Professor in Sri Krishna College of Engineering and Technology, Coimbatore She has about 5 years of Teaching Experience. 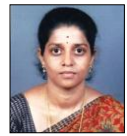